\documentclass[letterpaper, 10 pt, conference]{ieeeconf}  
\usepackage{etoolbox}
\makeatletter
\patchcmd{\@makecaption}
  {\scshape}
  {}
  {}
  {}
\makeatother

\IEEEoverridecommandlockouts                              

\usepackage{float}
\usepackage{graphicx} 
\usepackage{amsmath} 

\usepackage{soul,color}
\soulregister\cite7
\soulregister\ref7
\soulregister\pageref7

\usepackage{gensymb}

\title{\LARGE \bf
FLIVVER: Fly Lobula Inspired Visual Velocity Estimation \& Ranging
}

\author{Bryson Lingenfelter*\thanks{*Department of Computer Science and Engineering., University of Nevada, Reno. Email: blingenfelter@nevada.unr.edu}, Arunava Nag$^\dagger$\thanks{$^\dagger$Department of Mechanical Engineering., University of Nevada, Reno. Email: arunava\_nag@nevada.unr.edu, fvanbreugel@unr.edu}, and Floris van Breugel$^\dagger$
}

\begin{document}

\maketitle
\thispagestyle{empty}
\pagestyle{empty}

\begin{abstract}
The mechanism by which a tiny insect or insect-sized robot could estimate its absolute velocity and distance to nearby objects remains unknown. However, this ability is critical for behaviors that require estimating wind direction during flight, such as odor-plume tracking. Neuroscience and behavior studies with insects have shown that they rely on the perception of image motion, or optic flow, to estimate \textit{relative} motion, equivalent to a ratio of their velocity and distance to objects in the world. The key open challenge is therefore to decouple these two states from a single measurement of their ratio. Although modern SLAM (Simultaneous Localization and Mapping) methods provide a solution to this problem for robotic systems, these methods typically rely on computations that insects likely cannot perform, such as simultaneously tracking multiple individual visual features, remembering a 3D map of the world, and solving nonlinear optimization problems using iterative algorithms. Here we present a novel algorithm, FLIVVER\footnote{Flivver: (n) a cheap car or aircraft, especially one in bad condition}, which combines the geometry of dynamic forward motion with inspiration from insect visual processing to \textit{directly} estimate absolute ground velocity from a combination of optic flow and acceleration information. Our algorithm provides a clear hypothesis for how insects might estimate absolute velocity, and also provides a theoretical framework for designing fast analog circuitry for efficient state estimation, which could be applied to insect-sized robots.
\end{abstract}

\section{Introduction}

To navigate complex environments, robots and animals require accurate information about their current state relative to their surroundings. Two key components of that state are their absolute ground velocity and the distance to nearby objects. Humans, as well as other similarly scaled animals or robots, can estimate the distance to an object based on the relative position of that object on the two retinas. This mechanism is referred to as binocular parallax. Changes of these distance measurements over time provide an estimate of velocity. For smaller animals such as insects and insect-sized robots \cite{Ma2013, Floreano2015}, however, the error in stereo-based depth and velocity estimates is exponentially larger \cite{van_Breugel_2014, Verri1986}. 
In this paper we present a bio-inspired and bio-plausible algorithm for estimating velocity directly from image motion and acceleration information. 

We begin with a brief review of insect motion-vision. Over the past 30 years, the visual system of insects, including blow flies (\emph{Calliphora}), fruit flies (\emph{Drosophila}), locusts, and others has received a great deal of attention from neuroscientists. These studies have discovered a variety of neurons that respond to apparent image motion, also known as optic flow \cite{Malcolm1951}. Mathematically, optic flow is the image angular velocity $\dot{\alpha}$, defined for each receptor angle, $\alpha$, across the eye’s retina. Insects calculate optic flow using a feature-agnostic neural circuit that implements a ``delay-and-correlate" calculation between pairs of receptors \cite{Hassenstein1956,Yang2018,Borst2019}. These individual optic flow measurements are then pooled together by myriad neurons. The most celebrated of these are located in the lobula plate, and integrate optic flow across the entire field of view to respond to rotations about specific axes \cite{Krapp1996, Schnell2010, Suver2016}. Other related wide field cells are selective for looming stimuli \cite{Klapoetke2017}. In \emph{Calliphora}, directionally sensitive cells with smaller receptive fields ($\sim$40\degree) have also been described, referred to as FD (figure detection) cells \cite{Egelhaaf1985}. Finally, a recently described class of cells in \emph{Drosophila} referred to as the lobula columnar cells have smaller receptive fields and are sensitive expanding stimuli and motion of small objects \cite{Wu2016}. Still other neurons remain to be fully characterized \cite{Wei2019}. The diverse optic flow sensitive neurons described in insects can broadly be thought of as matched-filters for specific types of motion with different spatial receptive fields \cite{Franz2000}. Matched filtered optic flow has, previously, been used for obstacle avoidance and control in small quad-rotors \cite{Conroy2009}. In this paper, we take loose inspiration from these observations by using matched-filters and spatial-pooling of optic flow to solve an otherwise ill-conditioned mathematical problem.  

Optic flow is geometrically proportional to the ratio of velocity and distance (see Section \ref{sec:math}). Drawing on a nonlinear observability analysis, it can be shown that velocity and distance can be extracted from a sequence of measurements of their ratios, but only when a measurable and non-zero acceleration is applied \cite{van_Breugel_2014}. Although possible, developing algorithms to solve this problem that are robust and efficient remains an active area of research in computer vision and robotics. 

\section{Related Work}

Algorithms designed to solve the state estimation problem are generally referred to as SLAM (Simultaneous Localization and Mapping) algorithms. Developing SLAM algorithms that rely on a combination of visual and inertial information (VI-SLAM) has been of particular interest, because cameras and inertial measurement units (IMUs) are cheap and ubiquitous. The simplest SLAM algorithms for resolving scale ambiguity require two cameras, and rely on the known distance between the cameras to resolve the metric scaling of estimates (for example, \cite{ Ma2019}). With a single camera, IMU data is required to estimate the scale factor.

Monocular visual-inertial (VI) SLAM has recently received significant attention (for a concise review see \cite{Huang2019}). Although some efforts have been made to use optic flow for VI-SLAM \cite{Ummenhofer_2017_CVPR}, this approach relies on a neural network trained on specific images. As such, the scale factor does not reliably generalize to other environments. Most other approaches rely on tracking individual visual features across frames \cite{mur2017visual}\cite{von2018direct}. Because of the computational cost of tracking features, most real-time systems use sparse collections of visual features \cite{Engel2018}. Even when tracking individual features, however, the problem is quite challenging. One solution is to use a nonlinear Kalman Filter (e.g. an Extended Kalman Filter) to fuse data from the visual features and the IMU to estimate the scale factor \cite{Nutzi2011}. This approach, however, requires the integration of sensor measurements over the course of 10-15 seconds before the algorithm converges and proper SLAM can begin. With the growing interest in the field of monocular SLAM, Stumberg et. al \cite{von2018direct} took an approach of dynamic configuration of the scaling factor. Here, an initial arbitrary scale was assumed instead of delaying the initialization required to observe all the parameters. Although this approach allows initial state estimates to be made immediately, the algorithm still requires considerable time for the iterative method to converge to achieve an optimum scale. 

In general terms, most existing monocular SLAM algorithms rely on nonlinear joint optimizations of the visual and inertial measurements across sliding windows \cite{Leutenegger2014}. These computations require remembering past key frames, building and remembering a (sparse) 3D map of the world, and iteratively solving complex optimizations that are not guaranteed to converge using, for example, Gauss-Newton methods or Extended Kalman filters. By contrast, the FLIVVER algorithm directly computes absolute velocity without the need for memory beyond a few frames.

\section{Geometry of Forward Motion}
\label{sec:math}
\begin{figure}[t]
\centering
\includegraphics[width=.40\textwidth]{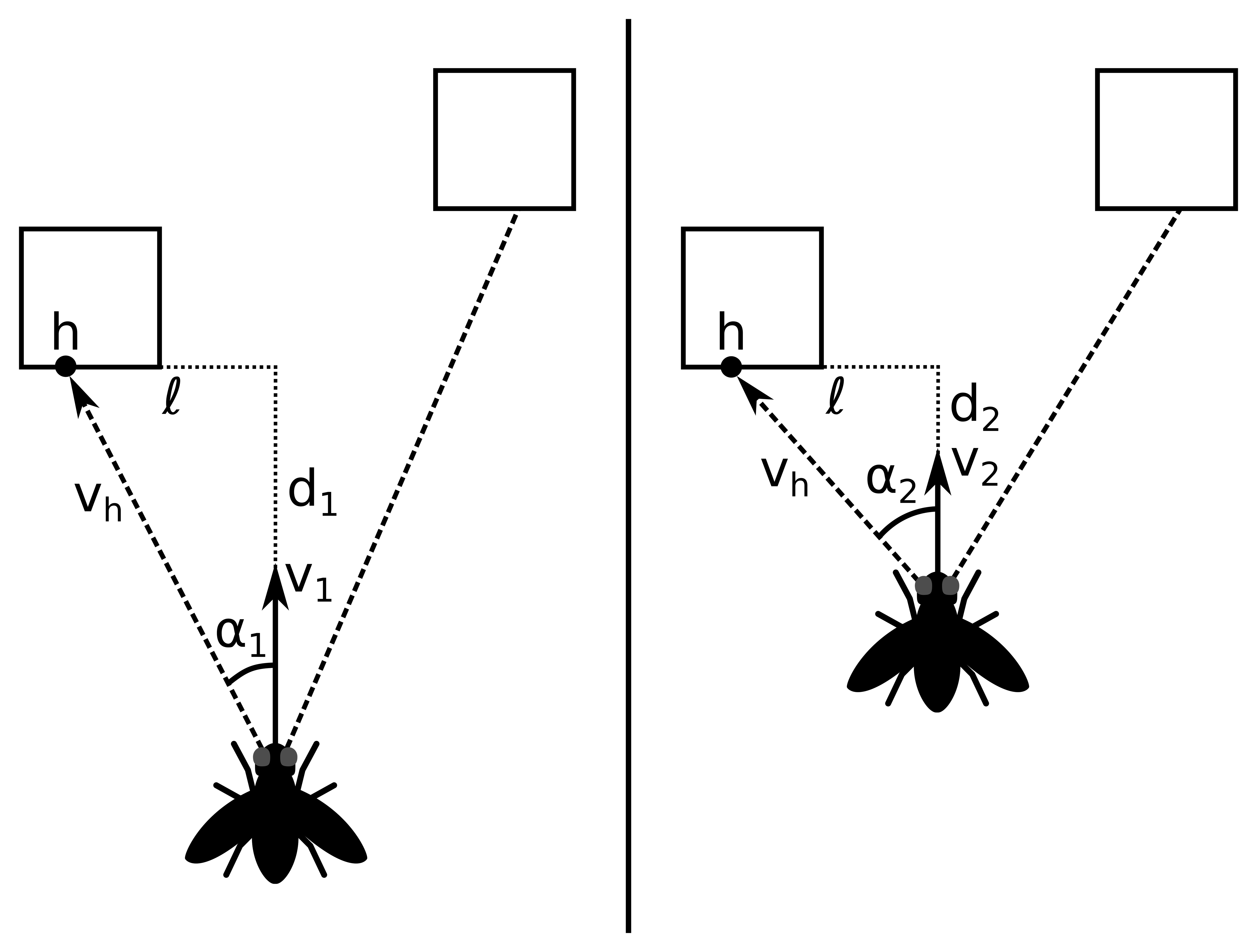}
\caption{Change in relationship to scene geometry over time during forward motion.}
\label{fig:flyflow}
\end{figure}

Although optic flow algorithms provide an estimate at each pixel, for visualization we consider only a single point. This is illustrated in Figure \ref{fig:flyflow}. The relationship between the forward distance $d$, and the lateral distance $l$ to a visual feature can be written as
\begin{equation}
tan(\alpha) = \frac{l}{d},
\label{eqn:l_over_d}
\end{equation}

where $\alpha$ is the angle between the center of the viewpoint and the visual feature. We derive Eqn \ref{eqn:l_over_d} with respect to time using the chain rule to find $\dot{\alpha}$, the optic flow: 
\begin{equation}
\frac{\dot{\alpha}}{cos^2(\alpha)} = \frac{\frac{\partial}{\partial t}l}{d} - \frac{l\frac{\partial}{\partial t}d}{d^2}.
\end{equation}
Flies spend the majority ($80\%$) of their time flying in straight lines \cite{vanBreugel2012}, which simplifies their visual experience by limiting it to pure forward translation. Flies are able to keep a straight heading using mechanosensory structures called halteres, which are thought to act as gyroscopes \cite{Pringle1948,Dickerson2019}. Inspired by this behavioral tendency to fly straight, our paper exclusively considers straight forward motion, though we discuss future extensions in Section \ref{sec:discussion}. Assuming perfectly straight motion, $\frac{\partial}{\partial t} l=0$, $l=d tan(\alpha)$, and $-\frac{\partial}{\partial t} d$ is simply the forward velocity, $v$. Plugging these values in and solving for $\dot{\alpha}$, we get the following relationship, 
\begin{equation}
\dot{\alpha} = \frac{-v}{d}cos(\alpha)sin(\alpha).
\end{equation}
If neither $v$ nor $d$ is known, optic flow alone does not give enough information to compute either. However, by taking the time derivative of $\frac{v}{d}$, the equation can be rewritten in a way that separates $v$ and $d$:

\begin{equation}
    \frac{\partial}{\partial t}\frac{(-v)}{d} = \frac{\frac{\partial}{\partial t}(-v)}{d}-\frac{(-v)^2}{d^2} = \frac{-a}{d}-\Big(\frac{v}{d}\Big)^2,
\end{equation}
where $a$ is the forward acceleration. This equation can now be rewritten as the following,

\begin{equation}
d = \frac{-a}{\big(\frac{v}{d}\big)^2-\frac{\partial}{\partial t}\frac{v}{d}},
\label{eqn:d_1}
\end{equation}

which we simplify by introducing a new variable, 
\begin{equation}
    r=\frac{v}{d}, 
    \label{eqn:r}
\end{equation}
such that Eqn \ref{eqn:d_1} becomes\footnote{Note that we define velocity in the forward direction here, whereas a previous publication defined $-v$ as the forward direction resulting in all three terms being positive}, 
\begin{equation}
d = \frac{-a}{r^2-\dot{r}}.
\label{eqn:d_2}
\end{equation}

Thus, so long as acceleration is non-zero, the distance to a visual feature in the forward direction can be estimated using measurements of acceleration, optic flow, and the derivative of optic flow. By combining Eqn \ref{eqn:d_2} with the relationship $d=v/r$, we can also find the equivalent relationship for estimating velocity,
\begin{equation}
v = \frac{-ar}{r^2-\dot{r}}.
\label{eqn:v_1}
\end{equation}
Note that Eqn \ref{eqn:v_1} is defined for each $\alpha$ in the visual field, corresponding to a receptor in the biological case, and to a pixel in the mechanical case. The velocity and acceleration, $v$ and $a$, however, are defined in the forward direction, and should therefore be identical for each pixel. We can take advantage of this by estimating $v$ for each pixel then computing the average to improve the overall estimate. 

Although these mathematical relationships show that it is theoretically possible to estimate velocity and distance, in practice several issues arise. First, the calculations are extremely sensitive to $\dot{r}$, which is difficult to estimate given noisy optic flow measurements. Second, the geometry is defined based on the image motion of an individual visual feature, not the motion seen by a given pixel. Thus, calculating $\dot{r}$ requires following features from one moment to the next, which is computationally expensive to do for a large number of features. If instead we attempt to calculate $\dot{r}$ on a per-pixel basis, new objects that come into, or leave, the field of view will dramatically change $\dot{r}$. Third, close to $\alpha=0$, Equations \ref{eqn:r}, \ref{eqn:d_2}, and \ref{eqn:v_1},  approach $\frac{0}{0}$, which causes the estimates to be unreliable in real world implementations. In the next section we draw on inspiration from a fly's visual processing to overcome these three issues.

\section{Bio-inspiration}
\label{sec:bio}

\begin{figure}[h]
\centering
\includegraphics[width=0.45\textwidth]{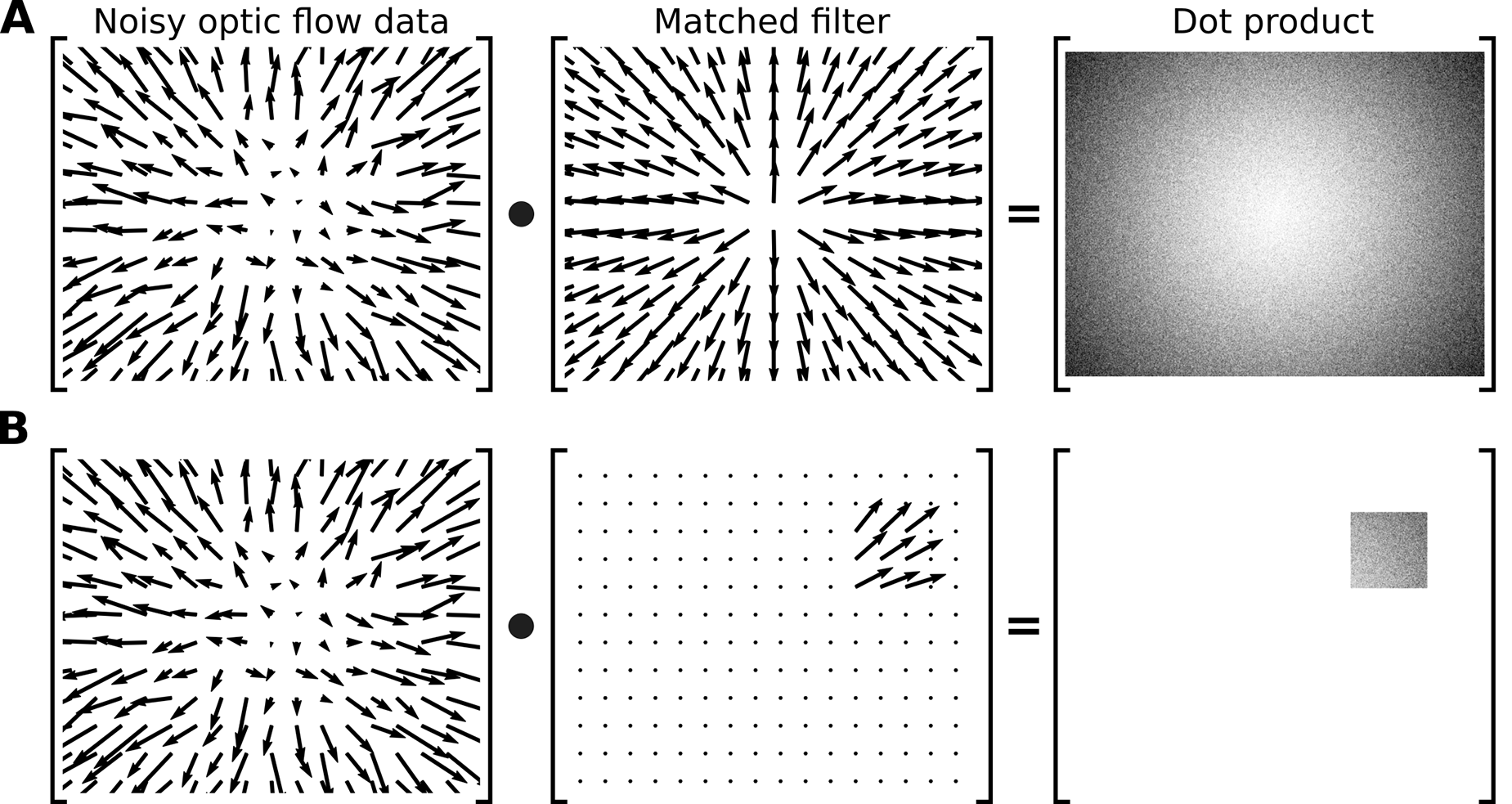}
\caption{Matched filtering can reduce noise when the expected flow profile is known. In this figure, noisy optic flow from forward motion is filtered according to the expected flow created by forward motion. (A) Shows the result for a full field filter, whereas (B) is for a smaller receptive field.}
\label{fig:match_filtering}
\end{figure}

To circumvent the practical issues that arise in implementing Eqn \ref{eqn:v_1}, we take inspiration from what is known about the visual processing circuitry in insects. In particular, we note that (a) the majority of fly visual processing circuitry throughout the lobula and lobula plate operate as matched filters, in that they are sensitive to particular kinds of motion, and (b) these neurons pool information across receptive fields that range from $\sim20-200^\circ$. Both operations can dramatically reduce noise in the measurements of $r$, which helps produce a better estimate of $\dot{r}$. Matched filtering reduces noise by only using optic flow information that matches what is expected based on the type of motion. For example, if the motion is straight forward, the optic flow field must be radiating out from the center. Any deviations from this are by definition noise and can be ignored (Fig. \ref{fig:match_filtering}). Spatial pooling further helps to reduce noise. Mathematically we can describe the spatial pooling by rewriting Eqn \ref{eqn:v_1} as follows,

\begin{equation}
\overline{\bigg(\frac{v}{a}\bigg)_{f_k}} = \overline{\Bigg(\frac{-r}{r^2-\dot{r}}\Bigg)_{f_k}},
\label{eqn:v_3}
\end{equation}

where $f_k$ represents a particular receptive field (region of interest) over which the mean is calculated. Rewriting this expression to the following equivalent,

\begin{equation}
\overline{v_{f_k}} = a\Bigg(\frac{-\overline{r_{f_k}}}{\overline{r_{f_k}^2}-\dot{\overline{r_{f_k}}}}\Bigg),
\label{eqn:v_4}
\end{equation}

where $a$ is the measured acceleration, allows us to spatially pool $\overline{r_{f_k}}$ before calculating $\dot{\overline{r_{f_k}}}$, reducing the effect of measurement noise on the derivative. Note, however, that $\overline{r_{f_k}^2} \neq {(\overline{r_{f_k}})^2}$. We can efficiently combine both the spatial pooling and matched filtering operations into a single step by taking the dot product between $r$, and $f_k$ and summing the resulting 2D matrix. This is equivalent to what small field optic flow sensitive neurons in a flies lobula accomplish. Repeating this calculation for many different receptive fields yields a collection of velocity estimates which can then be averaged to find a more accurate velocity estimate.  

\section{Dataset}
\label{sec:dataset}

\begin{figure}[!htb]
\centering
\includegraphics[width=.45\textwidth]{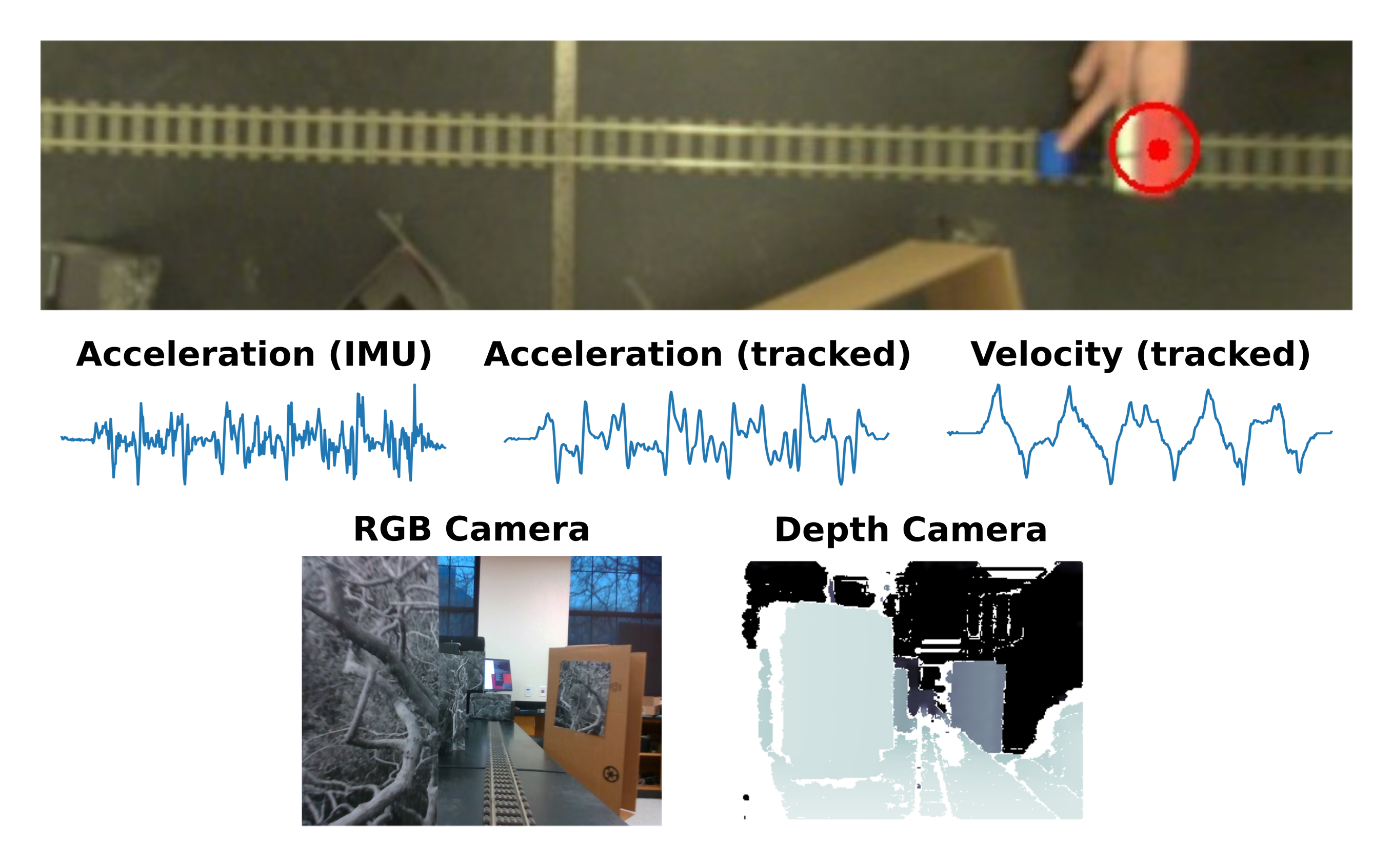}
\caption{Setup used for tracking ground-truth velocity.}
\label{fig:veltracking}
\end{figure}

We test our method using an Intel RealSense D435i depth camera. The camera provides monocular RGB images, depth images, accelerometer data, and gyroscope data. The depth camera is moved horizontally on a track surrounded by several objects. To collect ground-truth translation data for evaluation, we track the velocity of the depth camera using a second overhead camera. Our data collection setup is shown in Figure \ref{fig:veltracking}.

RGB and depth data are recorded at 30 frames per second, and odometry data is recorded at 250 frames per second. We average the odometry data such that there are 30 readings per second to be consistent with the camera data. To collect the dataset, the camera was moved forward and backward on a track for 80 seconds with varying acceleration.

\section{FLIVVER Implementation}
\label{sec:implementation}

\begin{figure*}[ht]
\centering
\includegraphics[width=1\textwidth]{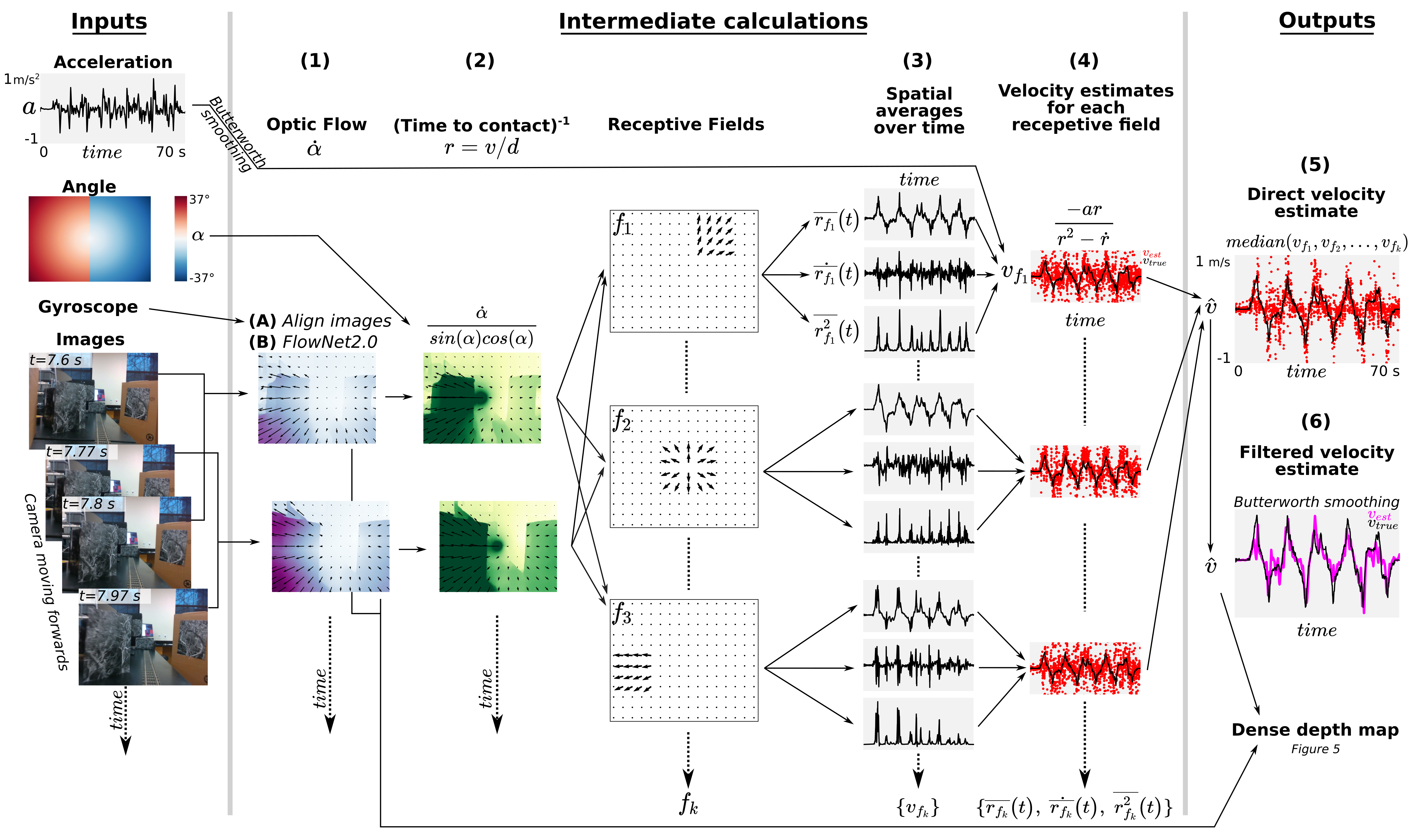}
\caption{Summary the FLIVVER algorithm for directly estimating velocity from optic flow and acceleration. Figure designed using FigureFirst \cite{Lindsay2017}. }
\label{fig:velocity_estimates}
\end{figure*}

\subsection{Estimating Optic Flow}
\label{sec:optic_flow}

Optic flow ($\dot{\alpha}$), the angular velocity at each pixel in the image, can be estimated using a variety of dedicated VLSI hardware systems \cite{Harrison2005, Mahalingam2010, Barrows2000}, and computer vision algorithms including Lucas Kanade \cite{Lucas1981, Baker2004}, and the more recent deep neural network, FlowNet 2.0 \cite{flownet2}. To compute optic flow, we use NVIDIA's Pytorch implementation of the FlowNet 2.0 model \cite{flownet2-pytorch} with the pre-trained model provided by the FlowNet 2.0 authors. Computer vision techniques for optic flow estimation give results in units of $pixels/frame$, which we convert to $radians/second$ using the camera's field of view and image resolution. 

30 frames per second of RGB data is useful for obtaining many samples to construct estimates from, but is problematic for calculating optic flow which is far more inaccurate when motion is very small. To deal with this, we stagger flow calculations as shown in Figure \ref{fig:velocity_estimates}. Flow at each frame is calculated as the flow between that frame and the flow six frames forward. This provides us a flow estimate every $\frac{1}{30}$ seconds, even though each individual flow calculation is using frames $\frac{1}{5}$ seconds apart.

\begin{figure*}[ht]
\centering
\includegraphics[width=1\textwidth]{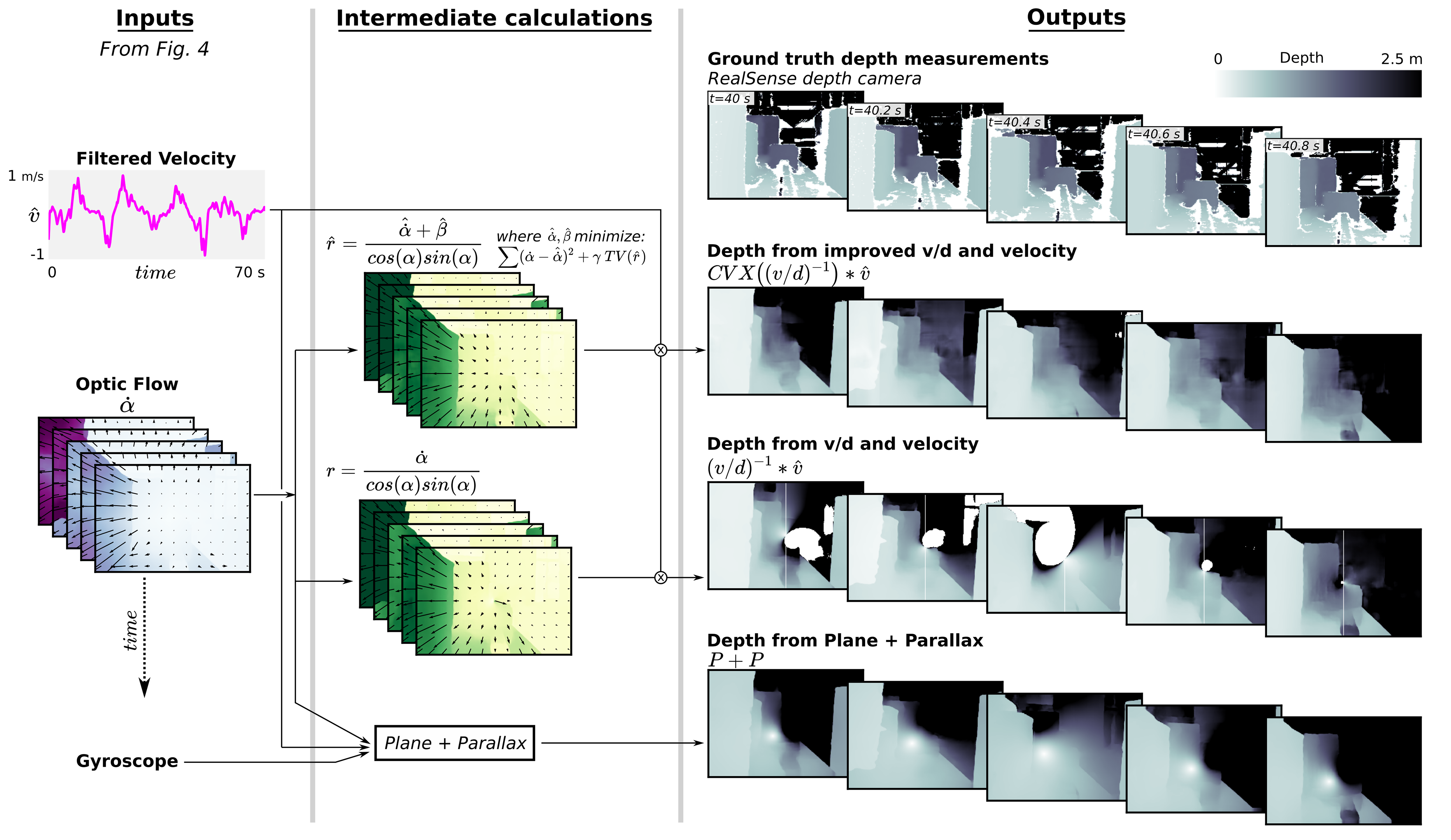}
\caption{Comparison of algorithms for estimating depth directly from optic flow, and the velocity estimates provided by the FLIVVER algorithm described in Fig. \ref{fig:velocity_estimates}. Note that the $CVX\big((v/d)^{-1}\big)*\hat{v}$ approach provides significantly better estimates, at the cost of the computation time. The image sequence chosen here is for positive velocity and negative acceleration.}
\label{fig:depth_estimates}
\end{figure*}

\subsection{Estimating Velocity Over Distance}
\label{sec:v_over_d}

Pure forward translation places substantial constraints on the flow field; if the scene is rigid, computation of optic flow is reduced to a 1D search problem \cite{irani2002direct},\cite{wulff2017optical}. Assuming strictly forward motion, flow should be directed out of the center of the image. We ensure this constraint holds by aligning images using gyroscope data to remove rotation. We then multiply the output of FlowNet by the full matched filter shown in Figure \ref{fig:match_filtering}, restricting the direction of the flow. Finally, we divide each pixel by $cos(\alpha)sin(\alpha)$ to obtain an estimate of $\frac{v}{d}$. This is done twice, once with horizontal angles and the horizontal component of flow and once with vertical angles and the vertical component of flow. Our final estimate of $\frac{v}{d}$ is the average of these two.

The general flow of the algorithm for calculating dense $\frac{v}{d}$ maps is shown in Figure \ref{fig:velocity_estimates}. First, optic flow is computed using two images 200 milliseconds apart. Next, the flow is aligned using a matched filter. Finally, the aligned flow is divided by $cos(\alpha)sin(\alpha)$ to produce an estimate of $\frac{v}{d}$. The $\frac{v}{d}$ estimate is highly noisy in the center of the image, where angles are very small, but otherwise reasonably approximates ground-truth $\frac{v}{d}$.

\subsection{Estimating Velocity}
\label{sec:vel_est}

Because of the challenges associated with differentiating noisy measurements of $v/d$, we use the receptive fields approach to described in Section \ref{sec:bio} to spatially pool our noisy $v/d$ measurements. In our implementation we use $90$ matched filters, each with a field of view of $5^\circ \times 5^\circ$, spanning the entire field of view. For each receptive field, we calculate $\overline{r_{f_k}}$ and $\overline{r_{f_k}^2}$. To estimate $\dot{\overline{r_{f_k}}}$, we first apply a Butterworth filter to $\overline{r_{f_k}}$ and then take a staggered finite difference derivative, so each value of $\overline{r_{f_k}}$ is compared to the value of $\overline{r_{f_k}}$ 5 frames previously. This delay further helps reduce noise in the estimates. Together, these three terms and the acceleration provide a velocity estimate for each receptive field according to Eqn \ref{eqn:v_4}. We then find the median velocity estimate.  

The resulting velocity estimates are still quite noisy, so we smooth them using a Butterworth filter. Estimates from the filter are in terms of forward velocity, so if there there was some amount of rotation this velocity is not the total velocity of the camera. We correct this by multiplying the velocity by $cos(a)$, where $a$ is the rotation change between the two frames. In our dataset this term is small enough to be negligible. For all Butterworth filtering we used a first-order filter with $w_n=0.04N_f$ where $N_f$ is the Nyquist frequency. 


\subsection{Estimating Depth}


Because our method provides estimates of $\frac{v}{d}$ and $v$, it is simple to invert $\frac{v}{d}$ and multiply by $v$ to obtain depth estimates.
In practice, this approach causes issues near the epipole. If the camera is not moving perfectly straight, the epipole (where $\dot{\alpha}=0$) is not aligned with $\alpha=0$. This offset results in pixels where $v/d=\epsilon/0$ instead of exactly $0/0$, causing $v/d$ to be erroneously large. Additionally, any noise in $\dot{\alpha}$ near the epipole will have a similar effect. These issues are avoided in the velocity calculations through spatial pooling, but for creating a dense depth map that is accurate near the epipole, they must be addressed. 

Future work will focus on efficient, bio-inspired, and bio-plausible methods. In this paper we present a preliminary approach that works well, but is too computationally expensive to run in real time. Our approach is to find $\hat{\dot{\alpha}}$, an estimate for the optic flow that is better aligned with $\alpha$. To find $\hat{\dot{\alpha}}$, we minimize the following loss function,

\begin{equation}
L = \sum_{}^{} (\dot{\alpha} - \hat{\dot{\alpha}})^2 + \gamma \; \sum_{}^{}\Bigg(\frac{\partial}{\partial \alpha}\bigg(\frac{\hat{\dot{\alpha}}+\hat{\beta}}{cos(\alpha)sin(\alpha)}\bigg)\Bigg)^2,
\label{eqn:cvx}
\end{equation}

where $\hat{\beta}$ is a constant offset, and $\gamma$ is a tuning parameter. The first term in $L$ ensures that $\hat{\dot{\alpha}}$ will remain faithful to the measured $\alpha$. The second term penalizes spatial derivative of $v/d$, and is equivalent to penalizing the total variation of $v/d$, so we can rewrite Eqn \ref{eqn:cvx} as follows, 

\begin{equation}
L = \sum_{}^{} (\dot{\alpha} - \hat{\dot{\alpha}})^2 + \gamma \; TV\bigg(\frac{\hat{\dot{\alpha}}+\hat{\beta}}{cos(\alpha)sin(\alpha)}\bigg).
\label{eqn:cvx2}
\end{equation}

The tuning parameter, $\gamma$ determines the balance between the faithfulness of $\hat{\dot{\alpha}}$ and the smoothness of $v/d$. $L$ is convex, and can efficiently be solved using convex optimization tools (such as cvxpy \cite{cvxpy} and the MOSEK solver). 

Alternatively, we can compute depth directly from optic flow. Once we have computed velocity we can reconstruct camera rotation and translation using gyroscope data. This allows us to use traditional methods for computing depth from optic flow. The plane+parallax representation can be used to compute depth estimates if the scene is rigid \cite{li2019learning}. These estimates are ill-defined near the epipole, but otherwise reasonably approximate scene geometry. Depth maps for all methods are shown in Figure \ref{fig:depth_estimates}.

\section{Evaluation}
\subsection{Accuracy of Velocity and Depth Estimates}

\begin{figure}[!htb]
\centering
\includegraphics[width=.4\textwidth]{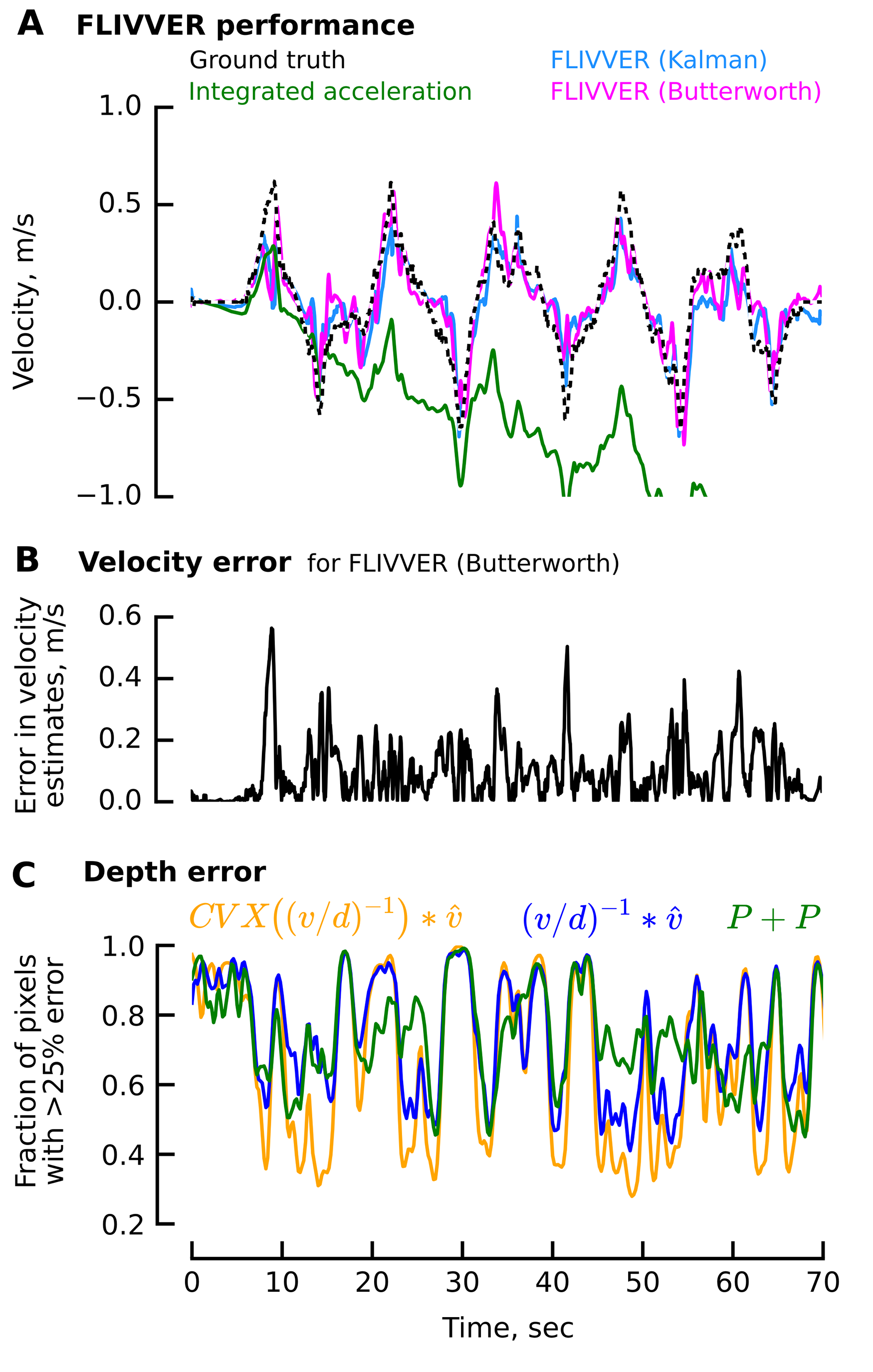}
\caption{FLIVVER performance. (A) Comparison between ground truth velocity (dashed black) and the FLIVVER algorithm summarized in Fig. \ref{fig:velocity_estimates} (magenta). Without true velocity estimates, integrating the acceleration accumulates large errors before long (green). Replacing the butterworth smoothing step with a constant acceleration Kalman smoother does not improve performance in a significant way. (B) FLIVVER velocity estimate errors as a function of time. (C) Error in depth estimates for the three methods shown in Fig. \ref{fig:depth_estimates}}
\label{fig:velocity_evaluation}
\end{figure}

We evaluate the FLIVVER algorithm by comparing the velocity estimates to ground truth data collected using an external tracking camera (see Section \ref{sec:dataset}). FLIVVER locks onto the correct absolute velocity in a few seconds, and achieves an $RMS$ error of $0.13$ (Fig. \ref{fig:velocity_evaluation}A-B). By comparison, naively integrating the acceleration results in the signal drifting away from the correct estimate within 5-10 seconds. For certain trajectories, a Kalman filter (cyan) may provide a better output than the simple Butterworth filter we use in the FLIVVER algorithm. However, for this dataset it does not improve the estimates ($RMSE=0.129$). Finally, we compare the three methods for estimating dense depth maps described in Fig. \ref{fig:depth_estimates} to the ground truth data from the RealSense camera (Fig. \ref{fig:velocity_evaluation}C). Generally, both the velocity and depth estimates are worst for high velocities, likely stemming from high levels of motion blur in the RGB images. 

\subsection{Computation Time}

A significant advantage of the FLIVVER approach is the potential for very fast computations times compared with feature-based tracking methods. This is because the majority of the computations can be done in parallel, and primarily consist of matrix multiplication or simple filtering. Our implementation prototype resulted in a total computation time of 110ms per frame. Almost all of this time is consumed by the computation of optic flow using FlowNet 2.0, as shown in Table \ref{tab:vel_comp}.

\begin{table}[h!]
  \begin{center}
    \caption{Velocity estimation computation times for a 448x640 pixel image stream. Flownet2 calculations were done on an NVIDIA GTX 1080 TI GPU. All other calculations were done on an Intel Xeon CPU E5-2620 @ 2.10GHz}
    \label{tab:vel_comp}
    \begin{tabular}{c|c|c|c}
      \textbf{Step} & \textbf{Description} & \textbf{Compute} & \textbf{Parallel}\\
      \textbf{ } & \textbf{ } & \textbf{time} & \textbf{threads}\\
      \hline
      1A & Image alignment (optional) & 13ms & no\\
      1B & Optic flow from Flownet2 & 93ms & no\\
      2 & (Time to contact)$^{-1}$ & $7\mu s$ & no\\
      3 & $\{\overline{r_{f_k}}, \;  \dot{\overline{r_{f_k}}}, \;  \overline{r^2_{f_k}}\}$ & 3.8ms & yes\\
      4 & Raw vel estimate & $0.3\mu s$ & yes\\
      5 & Median vel estimate & $8\mu s$ & no\\
      6 & Filtered vel estimate & $1\mu s$ & no\\
        & \textbf{Total} & 110ms &  \\
    \end{tabular}
  \end{center}
\end{table}

Directly computing baseline depth maps from velocity and $\frac{v}{d}$ only requires scalar inversion and multiplication, and can therefore be computed in microseconds. The convex optimization method, however, takes roughly 20 seconds per frame and is therefore not feasible for real-time applications.






\section{Discussion}
\label{sec:discussion}

In this paper we describe a novel algorithm, FLIVVER, for directly estimating forward velocity from a single camera using optic flow and acceleration. The goal of this research is two-fold: (a) to provide a concrete hypothesis for how insects might estimate their absolute ground velocity, and (b) to provide an efficient method for state estimation that could be implemented on size-power-weight constrained flying autonomous vehicles. Since FLIVVER provides a direct estimate of velocity without the need for algorithm convergence, it could also help reduce the convergence time for existing visual-inertial SLAM algorithms. The following subsections discuss limitations, and future focus areas.

\subsection{Limitations: Optic Flow}
The FLIVVER algorithm relies on high quality estimates of optic flow. In our implementation we used Flownet2, which does a good job of estimating optic flow even in texture-poor regions but is by far the slowest portion of our pipeline. Other, more efficient implementations may be less reliable. To reduce the effect of bad estimates resulting from low texture regions, the median calculation (step 5), could use weighted estimates, where the weighting is proportional to the quality of the optic flow estimate, based on the quality of the texture in a given window. 

\subsection{Limitations: Acceleration}
Any monocular SLAM algorithm requires non-zero accelerations to provide accurate absolute velocity estimates \cite{van_Breugel_2014}. FLIVVER is no different. Real-world implementations on robotic systems would likely benefit from the addition of a Kalman filter where the co-variance of the raw velocity estimate is inversely proportional to the magnitude of the acceleration. This would help improve velocity estimates for portions of the trajectory that have constant, and non-zero velocity. 

\subsection{Future Directions: Incorporating Rotation}
In the current implementation, we focused exclusively on straight forward motion. This is inspired by the observation that flies spend the majority of their time moving along straight paths, simplifying their visual experience. In the future, we will use a combination of gyroscopic measurements, and fly-inspired wide field matched filters to estimate rotation, allowing our algorithm to estimate both forward and rotational velocity. To estimate full 6-DOF velocity components will also require us to estimate lateral and vertical velocity. For this, we plan to develop bio-inspired algorithms for estimating the location of the epipole, rather than assuming it is in the center of the field of view.  

\subsection{Future Directions: Reducing Computation Times for Depth Estimates}
Our primary goal is to present an algorithm for estimating velocity. However, given that the combination of velocity estimates and optic flow make it possible to estimate dense depth maps, we present preliminary results for these as well. Estimating depth away from the epipole can be done very quickly. Close to the epipole, however, additional computation time is required to solve a convex optimization problem. Future work will focus on discovering a bio-plausible and efficient approach for these calculations, such as training an artificial neural network to accomplish the same task. 

\subsection{Future Directions: Analog Circuitry}
A major advantage of the FLIVVER algorithm for velocity estimation is that each computational step can, in theory, be implemented using analog circuitry, resulting in very fast state estimation. Although neural networks themselves can also, in principle, be implemented using analog circuits \cite{Du2019}, these circuits would require more custom hardware and significantly more components. Finally, the FLIVVER algorithm does not require any learning methods, and is therefore easier to implement in a general manner. Table \ref{tab:analog} summarizes the types of calculations involved in each step of FLIVVER's velocity estimation, and their analog equivalents.

\begin{table}[h!]
  \begin{center}
    \caption{Summary of the types of calculations FLIVVER uses (see Fig. \ref{fig:velocity_estimates}), and candidate analog equivalents.}
    \label{tab:analog}
    \begin{tabular}{c|c|c|c}
      \textbf{Step} & \textbf{Description} & \textbf{Current} & \textbf{Analog}\\
      \textbf{Step} & \textbf{Description} & \textbf{Approach} & \textbf{Equivalent}\\
      \hline
       & & & Analog VLSI \\
      1 & $\dot{\alpha}$ from Flownet2 & Neural net  & Optic Flow\\
       &  &  & \cite{Harrison2005, Mahalingam2010, Barrows2000}\\
      2 & (Time to contact)$^{-1}$ & Matrix Multiplication & \cite{Schlottmann2011}\\
      3 & $\{\overline{r_{f_k}}, \;  \dot{\overline{r_{f_k}}}, \;  \overline{r^2_{f_k}}\}$ & Matrix Multiplication & \cite{Schlottmann2011}\\
      4 & Raw vel estimate & Matrix Multiplication & \cite{Schlottmann2011}\\
      5 & Median vel estimate & Median Calculation & \cite{Dietz}\\
      6 & Filtered vel estimate & Butterworth filter & RLC circuit\\
    \end{tabular}
  \end{center}
\end{table}



\section*{Acknowledgements}
The authors thank James Hanna and Matteo Aureli for enlightening discussions leading to the selection of our algorithm's acronym.

\bibliographystyle{IEEEtran}
\bibliography{refs}

\end{document}